\newif\iffinal

\iffinal
	\documentclass[letterpaper, 10 pt, journal, twoside]{ieeetran}
\else
	\documentclass[letterpaper, 10 pt, conference]{ieeeconf} 
\fi

\IEEEoverridecommandlockouts        

\usepackage{cite}

\usepackage{amsmath,amssymb,amsfonts}
\usepackage{pifont}
\usepackage{graphicx}
\usepackage{textcomp}
\usepackage{xcolor}
\usepackage{color}													
\usepackage{xfrac}	
\usepackage{colortbl}
\usepackage{siunitx}
\usepackage[english]{babel}	
\usepackage{multirow}
\usepackage{hyperref}
\usepackage{blindtext}
\usepackage{mathtools}
\usepackage{hyperref}
\usepackage{tablefootnote}
\usepackage{algorithmic} 
\usepackage{textcomp} 
\usepackage{gensymb} 
\usepackage[ruled, linesnumbered]{algorithm2e}
\usepackage[caption=false, font=footnotesize]{subfig}
\usepackage{enumerate}
\usepackage{makecell}

\newcommand{\mr}[1]{\mathrm{#1}}
\newcommand{\bs}[1]{\boldsymbol{#1}}
\newcommand{\ks}[1]{$(\mr{CS})_{\mr{#1}}$}
\newif\ifhighlightchanges

\ifhighlightchanges
\newcommand{\highlightred}[1]{\textcolor{red}{#1}}
\else
\newcommand{\highlightred}[1]{#1}
\fi

\definecolor{klemmungfarbe}{RGB}{124,60,70}
\definecolor{kollisionfarbe}{RGB}{184,122,50}
\definecolor{imesorange}{RGB}{231,123,41}
\definecolor{xmark}{RGB}{186,0,0}
\definecolor{cmark}{RGB}{11,112,35}
\definecolor{gruen}{RGB}{0,152,70}
\definecolor{blau}{RGB}{0,0,255}
\definecolor{orange}{RGB}{255,110,0}
\definecolor{olivengruen}{RGB}{110, 117, 14}
\definecolor{gruenmarker}{RGB}{0, 158, 34}

\iffinal
	\title{Fast Contact Detection via Fusion of Joint and Inertial Sensors for Parallel Robots in Human-Robot Collaboration}

	\author{Aran Mohammad$^{1}$, Jan Piosik$^{1}$, Dustin Lehmann$^{2}$, Thomas Seel$^{1}$ and Moritz Schappler$^{1}$
		\thanks{Manuscript received: March 6, 2025; Accepted May 10, 2025.}
		\thanks{This paper was recommended for publication by Editor Clement Gosselin upon evaluation of the Associate Editor and Reviewers’ comments.
		This work was supported by the Deutsche Forschungsgemeinschaft~(DFG, German Research Foundation) under grant number 444769341.}
		\thanks{$^{1}$The authors are with the Institute of Mechatronic Systems, Leibniz University Hannover, 30823 Garbsen, Germany {\tt\footnotesize aran.mohammad@imes.uni-hannover.de}}%
		\thanks{$^{2}$The author is with the Technical University of Berlin, Control Systems Group, 10587 Berlin, Germany.}
		\thanks{Digital Object Identifier (DOI): see top of this page.}
	}
\else
	\title{\LARGE \bf
		Fast Contact Detection via Fusion of Joint and Inertial Sensors for Parallel Robots in Human-Robot Collaboration
	}

	\author{Aran Mohammad, Jan Piosik, Dustin Lehmann, Thomas Seel and Moritz Schappler
		\thanks{The authors are with the Institute of Mechatronic Systems, Leibniz University Hannover, 30823 Garbsen, Germany, and acknowledge the support of the German Research Foundation~(DFG) under grant number 444769341.
			{\tt\small aran.mohammad@imes.uni-hannover.de}}%
	}
\fi

\makeatletter
\newcommand{\removelatexerror}{\let\@latex@error\@gobble}
\makeatother

\newif\ifcopyright
\copyrighttrue
\begin{document}
	
	\ifcopyright
	{\LARGE IEEE Copyright Notice}
	\newline
	\fboxrule=0.4pt \fboxsep=3pt
	
	\fbox{\begin{minipage}{2\columnwidth}  
			Changes were made to this version by the publisher prior to publication. \\
			The final version of record is available at https://doi.org/10.1109/LRA.2025.3575326\\   
			Copyright (c) 2025 IEEE. Personal use of this material is permitted. For any other purposes, permission must be obtained from the IEEE by emailing pubs-permissions@ieee.org. \\
%
	\end{minipage}}
	\else
	\fi
	
	\graphicspath{{./graphics/}}
	\maketitle

	\begin{abstract}
		Fast contact detection is crucial for safe human-robot collaboration.
		Observers based on proprioceptive information can be used for contact detection but have first-order error dynamics, which results in delays.
		Sensor fusion based on inertial measurement units~(IMUs) consisting of accelerometers and gyroscopes is advantageous for reducing delays.
		The acceleration estimation enables the direct calculation of external forces.
		For serial robots, the installation of multiple accelerometers and gyroscopes is required for dynamics modeling since the joint coordinates are the minimal coordinates.
		Alternatively, parallel robots~(PRs) offer the potential to use only one IMU on the end-effector platform, which already presents the minimal coordinates of the PR.
		This work introduces a sensor-fusion method for contact detection using encoders and only one low-cost, consumer-grade IMU for a PR.
		The end-effector accelerations are estimated by an extended Kalman filter and incorporated into the dynamics to calculate external forces.
		In real-world experiments with a planar PR, we demonstrate that this approach reduces the detection duration by up to~50\% compared to a momentum observer and enables collision and clamping detection within~3--39\,ms.
	\end{abstract}
	\iffinal
		\begin{IEEEkeywords}
			Safety in human-robot interaction, parallel robots, sensor fusion
		\end{IEEEkeywords}
	\else
		\begin{keywords}
			Safety in human-robot interaction, parallel robots, sensor fusion
		\end{keywords}
	\fi
	
	\section{Introduction}
		\iffinal \IEEEPARstart{S}{afe} \else Safe \fi human-robot collaboration (HRC) is facilitated by limiting the kinetic energy via the robot's effective mass and velocity~\cite{InternationalOrganizationforStandardization.2016}.
		Lightweight serial robots reduce the energy by their design and operating at low velocities.
		Alternatively, the use of parallel robots (PRs) further \textit{reduces moving masses} due to the base-mounted drives~\cite{Merlet.2006}.
		As a result, the same energy limits can be maintained at \textit{higher velocities}.
		
		For both kinematic structures, contact with humans must be detected as quickly and robustly as possible.
		This can be realized via \textit{exteroceptive} or \textit{proprioceptive} information.
		Examples of the former include tactile skins~\cite{Dahiya.2013} or image-based approaches~\cite{Merckaert.2022}.
		\begin{figure}[t!]
			\vspace{0mm}			
			\centering
			\includegraphics[width=\columnwidth]{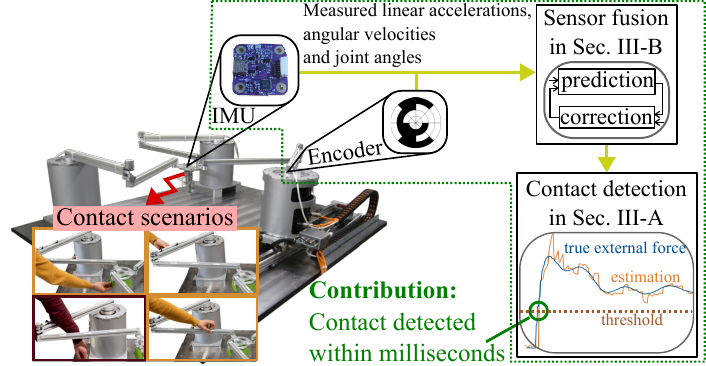}
			\caption{\textcolor{kollisionfarbe}{\textbf{Collision}} and \textcolor{klemmungfarbe}{\textbf{clamping}} contacts with a parallel robot are detected via a sensor fusion of one IMU and encoders. This approach enables contact detection within milliseconds by solving the dynamics equations~directly.}
			\label{fig:titelbild}
			\vspace{-2.5mm}
		\end{figure}
		The advantages of approaches with proprioceptive information are higher sampling rates, reduced time delays and reliability during dynamic contact scenarios.
		In~\cite{Popov.2017}, a neural network is trained with measured joint torques as inputs to perform \textit{data-driven} contact detection on a serial robot.
		The use of \textit{temporal} information improves the accuracy of data-driven contact detection by processing time series~\cite{Heo.2019,Park.2021}.
		In~\cite{Zhang.2021}, detection uncertainties are additionally considered through a series of sequential predictions.

		However, data-driven methods require extensive datasets for training in supervised learning or rely on prior knowledge of the trajectory in unsupervised learning to enable anomaly detection schemes.
		Consequently, they might exhibit limited accuracy in contact scenarios that are not within the training data set, in contrast to \textit{physics-based} approaches such as momentum observers (MOs), which demonstrate superior generalization capabilities.
		They infer external forces from the generalized momentum and require only sensors installed in the robot~\cite{Haddadin.2017}.
		The usage of an MO for contact detection is demonstrated for PRs in~\cite{Mohammad.2023}.
		In~\cite{Wahrburg.2015}, an MO is combined with a Kalman filter to consider modeling uncertainties and noise.
		Other methods include the use of an extended state estimator to observe the external force~\cite{Ren.2018}.
		
		However, the physical methods mentioned above share the problem that force estimation occurs with a \textit{phase delay}, which is caused by the \textit{observer's error dynamics}.
		Setting up faster dynamics conflicts with the robustness of the observer regarding noise and modeling inaccuracies.
		The \textit{direct approach}, where external forces are calculated using the dynamics equations, includes high-frequency components, unlike the MO, and enables faster contact detection~\cite{Birjandi.2020b}.
		But this approach requires information on the joint-angle \textit{acceleration}, which is not provided by the installed angle and current sensors.
		A double numerical differentiation of the angle measurement is \textit{inaccurate} due to noise, and an additional causal filtering leads to a \textit{phase delay}.
		
		Alternatively, \textit{sensor fusion} of kinematic information from an \textit{IMU} with the \textit{joint-angle sensors} can be used for faster and more accurate kinematic state estimation in robotics, as in the references on serial kinematic chains summarized in Table~\ref{tab:literatur}.
		In~\cite{Vihonen.2013}, one IMU is used per robot link.
		The joint velocity and acceleration are first determined using the measurement data from the gyroscope and the accelerometer.
		In addition, a Kalman filter estimates the joint angles and the bias of the gyroscope based on the encoder data.
		\begin{table}[t!]
			\vspace{3mm}
			\centering
			\caption{IMU-based estimation of kinematic states from SK and PK\tablefootnote{Serial/Parallel Kinematics (SK, PK), End Effector~(EE), Optimal~(Opt.)}}
			\label{tab:literatur}
			\vspace{-2.5mm}
			\begin{tabular}{c|c|c|c}
				Ref. & Type & Only one IMU? & Remarks \\  \hline
				\cite{Vihonen.2013} & \multirow{11}{*}{SK}& No (1 per link)  &  Gyroscope bias estimated \\ 
				\cite{MunozBarron.2015} &  & No (1 per link) &  --- \\ 
				\cite{McLean.2015}& & No (4 per link) &  Only simulation \\ 
				\cite{Rotella.2016} & & No (2 per link)  &  Gyroscope bias estimated \\ 
				\cite{Fennel.2022} &  & No (1 per link)  &  Sensor bias estimated\\ 
				\cite{Fennel.2023}& &No (less than 1 per link) &  Opt. number of sensors\\ 
				\cite{Birjandi.2023_ifac}& &No (1 per link)  & All joints in motion \\  
				\cite{Birjandi.2023}& &No (1 per link)  &  Covariance-adaptive\\ 
				\cite{Birjandi.2022}& & No (1 per link) &  Task-space control\\ 
				\cite{Birjandi.2020b} &  & No (1 per link)  & Force estimation \\ 
				\cite{Birjandi.2020}& & No (1 per link) & Model-adaptive \\ 
				\hline
				Ours & \textbf{PK} & \textbf{Yes (only 1 on EE)} & Force estimation
			\end{tabular}
			\vspace{-3mm}
		\end{table}
		By taking the sensor bias into account, the determination of the joint velocity is improved.
		Two cascaded Kalman filters are used in~\cite{MunozBarron.2015} for state estimation.
		The first fuses the encoder, gyroscope and acceleration sensors to estimate the joint angles.
		The second corrects the joint angle using the data from the gyroscope and estimates the joint-angle acceleration.
		Two unscented Kalman filters are used in a simulation of a serial robot in~\cite{McLean.2015} with four accelerometers per link, which allows a unique calculation of joint accelerations and angular rates.
		In~\cite{Rotella.2016}, joint velocity and acceleration are also determined via two IMUs for each link of a humanoid's leg, equivalently to a serial kinematic chain.
		Subsequently, two extended Kalman filters (EKFs) are used to improve the state estimation and, thereby, the feedback controller.
		In~\cite{Fennel.2022}, an EKF estimates joint coordinates, but additive offsets of the inertial sensors are also considered.
		The placement and number of accelerometers and gyroscopes in serial robots can be optimized to achieve equivalent performance regarding acceleration estimation in a simulation with fewer sensors than the robot's degrees of freedom~\cite{Fennel.2023}.
		The results of the kinematic states' estimation in~\cite{Birjandi.2023_ifac} show that fusing recursively one IMU per link of a 7-DoF robot in multi-joint motion allows an acceleration estimation with higher accuracy compared to double numerical differentiation.
		In~\cite{Birjandi.2023}, an adaptive EKF is demonstrated to consider implicit changes in sensor characteristics due to environmental factors.
		Task-space control laws based on velocity estimation can be deployed and extended to incorporate the planned trajectory formulated in task space by using differential kinematics~\cite{Birjandi.2022}.
		
		The direct approach to collision detection has already been successfully investigated in several studies involving serial robots.
		In~\cite{Birjandi.2020b}, kinematic state equations are formulated using joint angles and their time derivatives as states.
		The nonlinear output equation is the projection of the accelerations at the link coordinate system onto the translational accelerations at the IMU so that the states are corrected in an EKF depending on the measured accelerations.
		This enables force estimation~\cite{Birjandi.2020b}, as well as an adaptation of the robot dynamics models~\cite{Birjandi.2020}.

		The literature review shows that IMUs are suitable for acceleration estimation in robotics and for fast contact detection since they avoid the \textit{low-pass behavior of the classic MO}.
		However, these works only consider \textit{serial} robot kinematics whose minimum coordinates are the joint-space coordinates.
		The joint accelerations are therefore necessary for a clear description of the dynamics.
		Therefore, the basic premise is that multiple triaxial accelerometers and gyroscopes are installed on a serial robot, but this increases system complexity~\cite{Haddadin.2017}, as well as the deviation in the last links due to the recursive estimation and the error propagation~\cite{Birjandi.2023_ifac}.
		Even an observer-based optimization of the number and placement of sensors leads to a multiple-sensor setup for serial robots~\cite{Fennel.2023}.
		Only mounting one IMU on the serial robot's end effector is also not sufficient for joint acceleration estimation of commonly used \emph{redundant} robots.
		Mounting an IMU at the end effector of a \textit{six-DoF} serial robot and estimating the linear and angular end-effector acceleration is possible but not pursued in literature to the best of our knowledge.

		Similarly, a fusion of the encoders with \textit{only one IMU} is sufficient to describe the accelerations of \emph{parallel robots} since the \textit{end-effector platform coordinates are the minimal coordinates}.
		This attribute of PRs enables a minimal sensor-installation effort, requiring only one IMU.
		However, this advantage has not yet been sufficiently investigated in the context of fast and reliable contact detection for PRs, representing a research gap, which is marked in bold in Table~\ref{tab:literatur}.
		This motivates the contribution of this work depicted in Fig.~\ref{fig:titelbild} with the aim to address the described gap in the state of the art of IMU-based contact detection for PRs with the following three contributions:
		\begin{enumerate}[\bfseries{C}1]
		\item \label{contribution:1IMU}Fusing one IMU with the encoders is sufficient for estimating the accelerations of a planar PR's platform, which represents the minimal coordinates.
		\item \label{contribution:Determination_ExtForces}Estimated accelerations enable direct determination of external forces via the dynamics equations, resulting in contact detection within milliseconds.
		\item \label{contribution:Contact_Detection}Sensor fusion of one low-cost IMU and robot's encoders using an EKF is sufficient for better detection results compared to an MO --- shown within a simulation and on the real-world PR.
		\end{enumerate}
		The paper is organized as follows.
		Section~\ref{sec:preliminaries} defines the kinematics and dynamics modeling.
		The force determination and the sensor fusion are presented in Sec.~\ref{sec:direct}.
		Section~\ref{sec:simulation} describes a simulative evaluation of the sensor fusion and the force determination.
		In Sec.~\ref{sec:validation}, the experimental results are presented.
		Section~\ref{sec:conlusions} concludes this work.
	\section{Preliminaries} \label{sec:preliminaries}
		This section presents the kinematics~(\ref{ssec:kinematics}) and dynamics modeling~(\ref{ssec:dynamics}), followed by the disturbance observer~(\ref{ssec:observer}). 
		\begin{figure}[t!]
			\vspace{2.5mm}
			\centering
			\includegraphics[width=0.95\columnwidth]{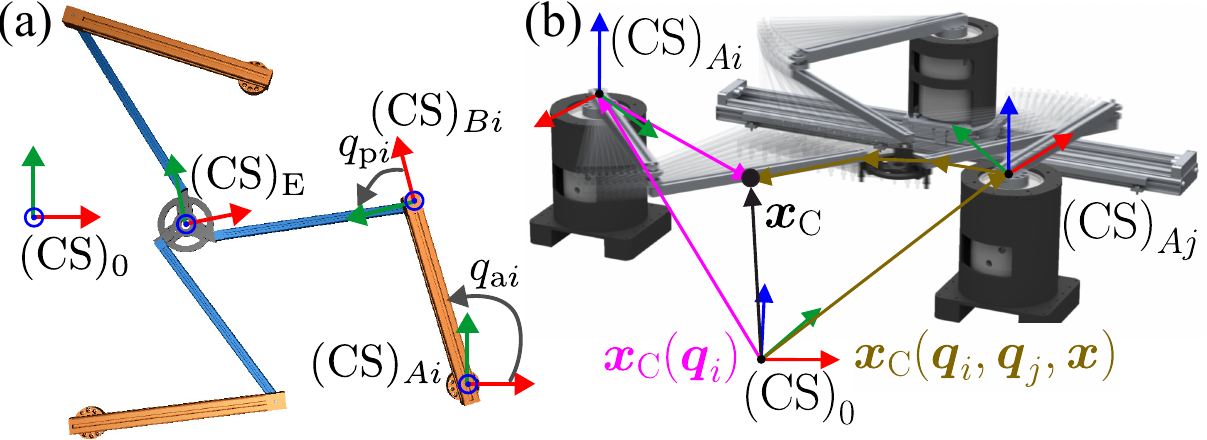}
			\caption{(a) The 3-\underline{R}RR parallel robot from~\cite{Mohammad.2023} with frames~$(\mr{CS})_\mr{E}$, $(\mr{CS})_{Ai}$, $(\mr{CS})_{Bi}$ at the end effector, active joint, passive joint and inertial frame~$(\mr{CS})_0$. (b) A contact at~$\bs{x}_\mr{C}$}
			\label{fig_3PRRR_real_skizze}
			\vspace{-3mm}
		\end{figure}
		\subsection{Kinematics} \label{ssec:kinematics}
			The modeling is described using a planar {3-\underline{R}RR} parallel robot\footnotemark~with the coordinate frames shown in Fig.~\ref{fig_3PRRR_real_skizze}(a).
			The PR has~$n{=}3$ end-effector degrees of freedom and~$m{=}3$ leg chains~\cite{Thanh.2012}.\footnotetext{The letter R denotes a revolute joint, and an underlining its actuation. The prismatic joint is kept constant and not considered in the modeling.}
			Operational-space coordinates (end-effector-platform pose\footnote{Corresponding quantities are expressed in~$(\mr{CS})_0$ unless specific frames are mentioned.}) and active joint's angles are respectively given by~$\bs{x}_{\mr{E}}^\mr{T}{=}[\bs{x}_{\mr{E,t}}^\mr{T},x_{\mr{E,r}}]{\in}\mathbb{R}^n$ and~$\bs{q}_\mr{a}{\in}\mathbb{R}^m$.
			The~$m_i{=}3$ joint angles of each leg chain in~$\bs{q}_i{\in}\mathbb{R}^{m_i}$ are stacked as~$\bs{q}^\mr{T}{=}[\bs{q}_1^\mr{T}, \bs{q}_2^\mr{T}, \bs{q}_3^\mr{T}] {\in} \mathbb{R}^{\highlightred{3m_i}}$.
			
			By closing vector loops~\cite{Merlet.2006}, kinematic constraints are constructed.
			The constraints' time derivatives give~$\dot{\bs{q}}{=}\bs{J}_{q,x}\dot{\bs{x}}_{\mr{E}}$ and~$\dot{\bs{x}}_{\mr{E}}{=}\bs{J}_{x,q_\mr{a}}\dot{\bs{q}}_\mr{a}$ with the Jacobian matrices\footnotemark~$\bs{J}_{q, x}{\in}\mathbb{R}^{3m\times n}$ and~$\bs{J}_{x, q_\mr{a}}{\in}\mathbb{R}^{n\times m}$.
			The notation~$\bs{J}_{a,b}$ represents the Jacobian matrix for the projection of the velocities of the coordinates in~$\bs{b}$ onto~$\bs{a}$.
			\footnotetext{For the sake of readability, dependencies on~$\bs{q}$ and~$\bs{x}_{\mr{E}}$ are omitted.}
			For the forward kinematics~$\bs{x}_{\mr{E}}{=}\bs{\mr{FK}}(\bs{q}_{\mr{a}})$, the Newton-Raphson approach is applied~\cite{Mohammad.2023}.
			
			The kinematics of an arbitrary (contact) point~$\mr{C}$ on the robot structure is modeled by formulating the contact coordinates~$\bs{x}_\mr{C}$ via joint angles~$\bs{q}$~\cite{Mohammad.2023}. 
			The pose~$\bs{x}_\mr{C}$ in Fig.~\ref{fig_3PRRR_real_skizze}(b) can be defined by closing vector loops.
			The time derivative results in~$\dot{\bs{x}}_\mr{C}{=}\bs{J}_{x_\mr{C},x}\dot{\bs{x}}_\mr{E}$ with the Jacobian~$\bs{J}_{x_\mr{C},x}$.
		
		\subsection{Dynamics} \label{ssec:dynamics}
			The Lagrangian equations of the second kind, along with the \textit{subsystem} and \textit{coordinate-partitioning} methods, formulate the dynamics in the operational space without the constraint forces~\cite{Thanh.2009}.
			The dynamics model\footnote{Generalized forces~$\bs{F}_x{\in}\mathbb{R}^n$ (including moments) in operational space} of the PR is
			\begin{equation} \label{eq_dyn}
				\bs{M}_x \ddot{\bs{x}}_{\mr{E}} + \bs{c}_x + \bs{g}_x + \bs{F}_{\mr{fr},x} = 	\bs{F}_{\mr{m},x} + \bs{F}_{\mr{ext},x}
			\end{equation} 
			with the notation~$(\cdot)_{x}$ representing a vector/matrix expressed in the operational-space coordinates~$\bs{x}_\mr{E}$.
			In~(\ref{eq_dyn}),~$\bs{M}_x$ denotes the inertia matrix,~$\bs{c}_x{=}\bs{C}_x\dot{\bs{x}}_{\mr{E}}$ the vector of the centrifugal and Coriolis effects,~$\bs{g}_x$ the gravitational terms,~$\bs{F}_{\mr{fr},x}$ the viscous and Coulomb friction components,~$\bs{F}_{\mr{m},x}$ the forces resulting from the motor torques and~$\bs{F}_{\mr{ext},x}$ the external forces.
			The forces~$\bs{F}_{\mr{m},x}$ are projected into the actuated-joint coordinates by the principle of virtual work as~$\bs{\tau}_\mr{m}{=}\bs{J}_{x,q_\mr{a}}^\mr{T}\bs{F}_{\mr{m},x}$.
			External forces~$\bs{F}_\mr{ext,link}$ at a link affect the end-effector platform via~$\bs{F}_{\mr{ext},x}{=}\bs{J}_{x_\mr{C},x}^\mr{T} \bs{F}_\mr{ext,link}$.
			A clamping contact causes two single link forces~$\bs{F}_\mr{ext,link1}$ and~$\bs{F}_\mr{ext,link2}$ at the locations~$x_{\mr{C}1}$ and~$x_{\mr{C}2}$~\cite{Mohammad.2023_Reaction}, which are projected to the end-effector platform coordinates with
			\begin{equation} \label{eq:jacobian_clamping}
				\begin{split}
					\bs{F}_{\mr{ext},x}=\bs{J}_{x_{\mr{C}1},x}^\mr{T} \bs{F}_\mr{ext,link1} + 	\bs{J}_{x_{\mr{C}2},x}^\mr{T} \bs{F}_\mr{ext,link2}.
				\end{split}
			\end{equation}
			
		\subsection{Generalized-Momentum Observer} \label{ssec:observer}
			The generalized momentum of a PR is set up in operational space with~$\bs{p}_x{=}\bs{M}_x \dot{\bs{x}}_\mr{E}$.
			The observer is based on the residual of this momentum.
			The time derivative of the residual is~$\sfrac{\mr{d}}{\mr{dt}} \hat{\bs{F}}_{\mr{ext},x} {=} \bs{K}_{\mr{o}} (\dot{\bs{p}}_x {-} \dot{\hat{\bs{p}}}_x )$ with~$\dot{\hat{\bs{p}}}_x{=}\dot{\hat{\bs{M}}}_x \dot{\bs{x}}_\mr{E} {+} \hat{\bs{M}}_x \ddot{\bs{x}}_\mr{E}$ and the observer-gain matrix~$\bs{K}_\mr{o}{=}\mr{diag}(k_{\mr{o},i}){>\bs{0}}$.
			By rearranging~(\ref{eq_dyn}) for~$\hat{\bs{M}}_x\ddot{\bs{x}}_\mr{E}$ and substituting it in the time integral of~$\sfrac{\mr{d}}{\mr{dt}} \hat{\bs{F}}_{\mr{ext},x}$, the generalized-momentum observer (MO)~\cite{Luca.2003} is constructed. 
			The resulting estimated external force is
			\begin{equation} \label{eq:mo}
					\hat{\bs{F}}_{\mr{ext},x} = \bs{K}_\mr{o} \left( \hat{\bs{M}}_x \dot{\bs{x}}_\mr{E} {-} \int_{0}^t 	(\bs{F}_{\mr{m},x} {-} \hat{\bs{\beta}}_x {+} \hat{\bs{F}}_{\mr{ext},x}) \mr{d}\tilde{t} \right),
			\end{equation}
			with~$\dot{\hat{\bs{M}}}_x{=}\hat{\bs{C}}_x^\mr{T}{+}\hat{\bs{C}}_x$~\cite{Haddadin.2017, Ott.2008} and~$\hat{\bs{\beta}}_x{=}\hat{\bs{g}}_x {+} \hat{\bs{F}}_{\mr{fr},x}{-}\hat{\bs{C}}_x^\mr{T} \dot{\bs{x}}_\mr{E}$.
			When assuming~$\hat{\bs{\beta}}_x {=} \bs{\beta}_x$, the MO's estimation exponentially converges to the external force in the platform coordinates with the linear and decoupled first-order error dynamics
			\begin{align}\label{eq:mo_error_dynamics}
				\bs{K}_\mr{o}^{-1} \dot{\hat{\bs{F}}}_{\mr{ext},x} + \hat{\bs{F}}_{\mr{ext},x}{=}\bs{F}_{\mr{ext},x}.
			\end{align}
			The parameterization of~$\bs{K}_\mr{o}$ influences the observer's performance and robustness.
			High values in~$\bs{K}_\mr{o}$ reduce the phase delay and improve the sensitivity for high-frequency oscillations, but they amplify the noise influence, too.
			In extreme cases, this leads to false-positive contact detections in an HRC scenario, resulting in less efficient robot operation. 
			Noise influence can be reduced by low entries in~$\bs{K}_\mr{o}$, but this leads to more extensive phase delays and finally to false-negative results.
			The following approach resolves this trade-off between high detection performance and robustness.
	\section{Sensor Fusion for Force Determination} \label{sec:direct}
		The main contributions, the direct method~(\ref{ssec:direkteMethode}) and the sensor fusion~(\ref{ssec:sensorfusion}) are introduced in this section.
		\subsection{Determination of External Forces} \label{ssec:direkteMethode}
\begin{figure}[t!]
				\vspace{2.5mm}
				\centering
				\includegraphics[width=0.95\columnwidth]{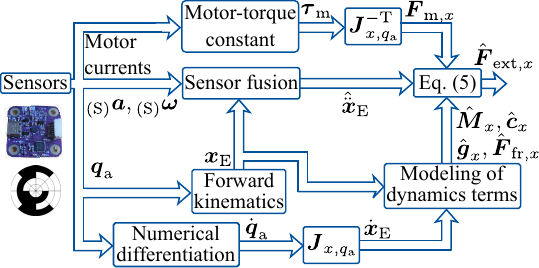}
				\caption{Flow chart of the introduced direct method with sensor fusion}
				\label{fig:umsetzung}
				\vspace{-3mm}
			\end{figure}

			Equation~\ref{eq_dyn} is rearranged to determine the external force
			\begin{equation} \label{eq:dyn_Fext}
				\bs{F}_{\mr{ext},x} = \bs{M}_x \ddot{\bs{x}}_{\mr{E}} + \bs{c}_x + \bs{g}_x + \bs{F}_{\mr{fr},x} - \bs{F}_{\mr{m},x}.
			\end{equation} 
			Hence, this method is called the \textit{direct method}.
			The process is depicted in Fig.~\ref{fig:umsetzung}.
			The dynamics parameters for the terms~$\bs{M}_x, \bs{c}_x$ and~$\bs{g}_{x}$ are identified using the method described in~\cite{Thanh.2009}.
			The friction losses~$\bs{F}_{\mr{fr},x}$ are approximated using Coulomb and viscous friction models.
			The active joints' torques~$\bs{\tau}_{\mr{m}}$ are determined via the measured motor current and the identified motor-torque constant.
			The force~$\bs{F}_{\mr{m},x} {=} \bs{J}_{x,q_{\mr{a}}}^{-\mr{T}}\bs{\tau}_{\mr{m}}$ is given by transformation into operational space.
			The end-effector pose and velocity are provided by applying the forward kinematics~$\bs{x}_{\mr{E}}{=}\bs{\mr{FK}}(\bs{q}_{\mr{a}})$ and differential kinematics~$\dot{\bs{x}}_{\mr{E}}{=}\bs{J}_{x,q_\mr{a}}\dot{\bs{q}}_\mr{a}$.
			A numerical determination of the end-effector acceleration is inaccurate due to the encoders' noisy measurements, as will be shown later.
			
			A Kalman-filter-based sensor fusion of an IMU and encoders is used to estimate~$\ddot{\bs{x}}_{\mr{E}}$.
			Hence, the method presented here is a direct determination of external forces using sensor fusion to estimate kinematic states.
			This has already been investigated for serial robots~\cite{Birjandi.2020b} but is novel for parallel kinematics.
			From the sensor-fusion outputs, only the estimated accelerations~$\hat{\ddot{\bs{x}}}_\mr{E}$~(and not the platform pose and velocities) are provided to~(\ref{eq:dyn_Fext}) due to the high accuracy of end-effector pose and velocity determination via the PR's forward kinematics~$\bs{x}_{\mr{E}}{=}\bs{\mr{FK}}(\bs{q}_{\mr{a}})$ and differential kinematics~$\dot{\bs{x}}_{\mr{E}}{=}\bs{J}_{x,q_\mr{a}}\dot{\bs{q}}_\mr{a}$.
			This approach avoids the occurrence of drifting and the associated risk of mis-detection caused by the time integration of acceleration.
		\subsection{Sensor Fusion for Estimation of Kinematic States} \label{ssec:sensorfusion}
			Simply using the measured three-dimensional accelerations and angular velocities from the IMU for force calculation in~(\ref{eq:dyn_Fext}) is insufficient since~(i) rotational accelerations of operational-space coordinates are not given and~(ii) the measurements are noisy and biased.
			These two limitations are addressed by fusing the IMU with the joint-angle encoders to estimate the kinematic states and finally determine external forces.

			The \highlightred{discrete-time} state-space model of the observer is
			\begin{equation}	\label{eq.nulla}
				\bs{x}_{k} {=} \bs{f}(\bs{x}_{k-1}){+}\bs{w}_{k-1}{=} \begin{bmatrix} \bs{I} & T \, \bs{I}  & \frac{T^2}{2} \, \bs{I} \\ \bs{0} & \bs{I} & T \, \bs{I} \\ \bs{0} & \bs{0} & \bs{I} \end{bmatrix}  \bs{x}_{k-1} {+} \bs{w}_{k-1},
			\end{equation}
			with $\bs{I}$ and $\bs{0}$ as the ${n{\times}n}$ identity and zero matrix, $T{\in}\mathbb{R}_{>0}$ as the sampling time, the end-effector pose, velocity and acceleration as states in~$\bs{x}_k{=}[\bs{x}_{\mr{E}}^\mr{T},\dot{\bs{x}}_{\mr{E}}^\mr{T},\ddot{\bs{x}}_{\mr{E}}^\mr{T} ]_k^\mr{T}{\in}\mathbb{R}^{3n}$ and~$\bs{w}_k {\in} \mathbb{R}^{3n}$ describing the process noise.
			The Euler angles are chosen for orientation representation due to their simplicity. 
			The approximation error of~(\ref{eq.nulla}) is assumed to be negligible due to the high sampling rate.
			The occurrence of the gimbal-lock configuration is beyond the scope of this work, as the majority of parallel robots can not reach tilting angles of~${\pm}\SI{90}{\deg}$ and the investigated planar parallel robot does not encounter the problem since it has only one orientation degree of freedom in the task space.
			
			The end-effector pose~$\bs{x}_{\mr{E}}$ via the forward kinematics based on encoders and the IMU mounted at the end-effector platform are used in the output equation 
			\begin{equation}\label{eq.mess}
				\bs{y}_k {=} \bs{g}(\bs{x}_{k}){+}\bs{v}_{k} {=} \begin{bmatrix} \bs{x}_{\mr{E}} \\
				_{(\mr{S})}\bs{\omega} (\bs{x}_{\mr{E}}, \dot{\bs{x}}_{\mr{E}}) \\
				_{(\mr{S})}\bs{a}_{\mr{S}} (\bs{x}_{\mr{E}}, \dot{\bs{x}}_{\mr{E}}, \ddot{\bs{x}}_{\mr{E}}) 		\end{bmatrix}_{k} {+}\bs{v}_{k}.
			\end{equation}
			From the IMU measurements, the acceleration~$_{(\mr{S})}\bs{a}_{\mr{S}}$ and angular velocity~$_{(\mr{S})}\bs{\omega}$ are obtained in the frame~\ks{S} fixed to the IMU.
			The measurement noise is described by~$\bs{v}_k {\in} \mathbb{R}^l$, with~$l$ as the output dimension.
			The IMU's acceleration can be decomposed into 
			\begin{align}\label{eq.messa}
				_{(\mr{S})}\bs{a}_{\mr{S}}{=} \, ^{\mr{S}}\bs{R}_{\mr{E}} \ _{(\mr{E})}  \bs{a}_{\mr{S}}{ =} & \, 	^{\mr{S}}\bs{R}_{\mr{E}} ( _{(\mr{E})}\ddot{\bs{x}}_{\mr{E,t}} {+} _{(\mr{E})}\dot{\bs{\omega}}_{\mr{E}} \times _{(\mr{E})}\bs{p}_{\mr{S}} \\ & {+} _{(\mr{E})}\bs{\omega}_{\mr{E}} {\times} ( _{(\mr{E})}\bs{\omega}_{\mr{E}} {\times} _{(\mr{E})}\bs{p}_{\mr{S}} ) {+} ^{\mr{E}}\bs{R}_{0}{} _{(\mr{0})}\bs{g} ),  \nonumber
			\end{align}
			where~$^{\mr{S}}\bs{R}_{\mr{E}}$, the rotation from the end-effector frame~\ks{E} to~\ks{S}, and~$_{(\mr{E})}\bs{p}_{\mr{S}}$, the IMU's position in~\ks{E}, are constant, as the IMU is fixed to the end effector.
			The sensors' acceleration~$_{(\mr{E})}  \bs{a}_{\mr{S}}$ is composed of the angular acceleration~${_{(\mr{E})}\dot{\bs{\omega}}_{\mr{E}}} {\times} _{(\mr{E})}\bs{p}_{\mr{S}}$, the centripetal acceleration~$_{(\mr{E})}\bs{\omega}_{\mr{E}} {\times} ( _{(\mr{E})}\bs{\omega}_{\mr{E}} {\times} _{(\mr{E})}\bs{p}_{\mr{S}} )$ and the gravitational-acceleration vector~${_{(\mr{0})}}\bs{g}$.
			The end-effector's absolute acceleration~$_{(\mr{E})}\ddot{\bs{x}}_{\mr{E,t}}$ results from the rotation~$^{\mr{E}}\bs{R}_{0}$ and the acceleration in the inertial frame~\ks{0}.			
			The angular velocity
			\begin{equation}	\label{eq.Hmatrix}
				_{(\mr{E})}\bs{\omega}_{\mr{E}}
				={^{\mr{E}}\bs{R}_{\mr{0}}} _{(0)}\bs{\omega}_{\mr{E}}
				= {^{\mr{E}}\bs{R}_{\mr{0}}} \bs{J}_\omega {}_{(0)}\dot{\bs{x}}_{\mr{E,r}}
			\end{equation}
			of the end effector results from the time derivatives~$_{(0)}\dot{\bs{x}}_{\mr{E,r}}$ of the Euler angles that are multiplied by the rotation $^{\mr{E}}\bs{R}_{\mr{0}}$ and matrix~$\bs{J}_\omega$\footnote{For a planar system, $\bs{J}_\omega$ is the identity matrix.}. 
			The output $_{(\mr{S})}\bs{\omega} {=} \, ^{\mr{S}}\bs{R}_{\mr{E}} \, _{(\mr{E})}  \bs{\omega}_{\mr{E}}$ of the gyroscope is obtained by rotating~(\ref{eq.Hmatrix}) into~\ks{S}.
			The IMU's mounting position~$_{(\mr{E})}\bs{p}_{\mr{S}}$ and the intrinsic Euler angles in~$\bs{\varphi}_{xyz}^\mr{S,E}$ for describing the rotation~$^{\mr{S}}\bs{R}_{\mr{E}}$ are identified via the optimization of
			\begin{align}\label{eq:cali}
				\min_{{_{(\mr{E})}\bs{p}_{\mr{S}}},\bs{\varphi}_{xyz}^\mr{S,E}} 
				\sum_{i=1}^{n_\mr{meas}} \left(
				\frac{||{_{(\mr{S})}\bs{\omega}_i}{-}{_{(\mr{S})}\hat{\bs{\omega}}_i}||_2^2} {\omega_{\max}^2}
				{+}
				\frac{||{_{(\mr{S})}\bs{a}_{\mr{S},i}}{-}{_{(\mr{S})}\hat{\bs{a}}_{\mr{S},i}}||_2^2} {a_{\max}^2}
				\right)
			\end{align}
			using~$n_\mr{meas}$ measurement samples.
			Maximum absolute values~$a_{\max}$ and~$\omega_{\max}$ of measured translational accelerations and angular velocities are used to compensate for the effect of different units and range values.
			
			Due to the nonlinearity in~(\ref{eq.mess}), an extended Kalman filter~(EKF) is chosen.
			\begin{figure}[t!]
				\removelatexerror
				\vspace{1.5mm}
				\begin{algorithm}[H] \nonumber
					\caption{Estimation by the EKF}
					\label{alg:estimate_ekf}
					{\small 
						\SetKwInOut{Input}{Input}
						\SetKwInOut{Output}{Output}
						\Input{$\hat{\bs{x}}_{k-1}$, $\bs{y}_k$, $\bs{P}_{\bs{x}_{k-1}}$, $\bs{Q}_\mr{ekf}$, $\bs{R}_\mr{ekf}$} 
						\Output{$\hat{\bs{x}}_k,\bs{P}_{\bs{x}_k}$}
						\SetKwFunction{FMain}{checkStatus}
						\SetKwProg{Fn}{Function}{:}{}
						\tcp{process update}
						$\bs{x}_k^{-} \gets \bs{f}(\hat{\bs{x}}_{k-1})$\tcp*{mapping by~$\bs{f}$}\
						$\bs{A}_k \gets \frac{\partial \bs{f}}{\partial \bs{x}}\bigg\rvert_{\hat{\bs{x}}_{k-1}}$\tcp*{linearization of~$\bs{f}$}\
						$\bs{P}_{\bs{x}_k}^{-} \gets \bs{A}_k \bs{P}_{\bs{x}_{k-1}} \bs{A}_k^\mr{T} +\bs{Q}_\mr{ekf};$\\
						\tcp{measurement update}
						$\bs{y}_k^{-} \gets \bs{g}(\bs{x}_k^{-})$\tcp*{mapping by~$\bs{g}$}\
						$\bs{C}_k \gets \frac{\partial \bs{g}}{\partial \bs{x}}\bigg\rvert_{\bs{x}_k^{-}}$\tcp*{linearization of~$\bs{g}$}\
						$\bs{K}_k{ \gets }\bs{P}_{\bs{x}_k}^{-}\bs{C}_k^\mr{T} (\bs{C}_k\bs{P}_{\bs{x}_k}^{-}\bs{C}_k^\mr{T}{+}\bs{R}_\mr{ekf})^{-1};$\\
						$\bs{P}_{\bs{x}_k}{\gets}\bs{P}_{\bs{x}_k}^{-}{-}\bs{K}_k\bs{C}_{k}\bs{P}_{\bs{x}_k}^{-}$\tcp*{state covariances}\
						%
						%
						$\hat{\bs{x}}_k \gets \bs{x}_k^{-}+\bs{K}_k(\bs{y}_k-\bs{y}_k^{-})$\tcp*{estimate states}\
						\KwRet $\hat{\bs{x}}_k,\bs{P}_{\bs{x}_k}$\;
					}
				\end{algorithm}
				\vspace{-4mm}
			\end{figure}
			The EKF's steps shown in Alg.~\ref{alg:estimate_ekf} consist of the process (lines~1--3) and measurement updates~{(lines 4--8)} with the estimated states~$\hat{\bs{x}}_{k-1}$, measured outputs~$\bs{y}_k$, state-covariance matrix~$\bs{P}_{\bs{x}_{k-1}}$, process-covariance matrix~$\bs{Q}_\mr{ekf}$ and measurement-covariance matrix~$\bs{R}_\mr{ekf}$ as inputs.
			The estimation accuracy depends on the covariances in the diagonal matrices~$\bs{Q}_\mr{ekf}$ and~$\bs{R}_\mr{ekf}$, which are assumed to be uncorrelated.
			
			The nonlinear system must be observable to estimate states using sensor fusion.
			Proving full observability of such a system is complex.
			However, it is possible to prove local observability instead~\cite{Khalil.2014}.
			By linearizing~(\ref{eq.mess}) in the operating point, observability can be verified using methods for linear systems.
			The Kalman criterion is checked for all operating points of the trajectories in the following chapters.
	\section{Simulative Experiments} \label{sec:simulation}
		The simulation environment~(\ref{ssec:simenv}) and the observer parameterization~(\ref{sec.auslegung}) are described. Finally, the results of the sensor fusion and the direct method are presented~(\ref{ssec:simres}).
		\subsection{Simulation Environment} \label{ssec:simenv}
			For the simulative analysis, a Simulink model with an embedded MuJoCo model~\cite{Todorov.2012} of the PR is used. 
			The active-joint angle measurements from the encoders~(Heidenhain \href{https://www.heidenhain.de/fileadmin/pdf/de/01_Produkte/Prospekte/PR_Messgeraete_fuer_elektrische_Antriebe_ID208922_de.pdf}{ECN1313}) are assumed to contain a mean-free white noise with a standard deviation of~$4{\cdot} 10^{-4}\SI{}{\deg}$ and a resolution of~$1.2 {\cdot} 10^{-5}\SI{}{\deg}$.
			The output~$\bs{x}_{\mr{E}}$ is then determined using forward kinematics.
			The IMU is mounted on the end effector in the MuJoCo simulation.
			The IMU~(\href{https://www.bosch-sensortec.com/media/boschsensortec/downloads/datasheets/bst-bmi160-ds000.pdf}{BMI160} from Bosch) measurements are initially assigned a mean-free white noise of~$180{\cdot}10^{-6}\SI{}{ \textit{g}/ \hertz}$ for the accelerometer and~$\SI{0.007}{\deg/ \second / \hertz}$ for the gyroscope, as well as a bias of~$\SI{\pm 0.04}{\textit{g}}$ and~$\SI{\pm 3}{\deg / \second}$ respectively.
			The accelerometer has a resolution of~$2.99{\cdot}10^{-4}\SI{}{\meter\per\second^2}$ and the gyroscope of~$1.91{\cdot}10^{-3}\SI{}{\deg\per\second}$.
		\subsection{Observer Parameterization} \label{sec.auslegung}
			The covariance matrix is initialized with~$\bs{P}_0 {=} 0.1 {\cdot} \bs{I}_{3n}$.
			The variances of the IMU's outputs for~$\bs{R}_\mr{ekf}$ are taken from the sensor's datasheets mentioned in Sec.~\ref{ssec:simenv}.
			For simplicity, the variance of the platform pose $\bs{x}_\mr{E}$ is determined heuristically and not adaptively via the Jacobian matrices and the encoder noise.
			The diagonal entries of~$\bs{Q}_\mr{ekf}$ are determined heuristically.
			The resulting matrices are 
			\begin{subequations}\label{eq:ekf_setting}
				\begin{align}
					\bs{R}_{\mr{ekf}} &{=} \mr{diag} (1.2 {\cdot} 10^{-1} \bs{I}_n, \, 1.6 {\cdot} 10^{-3} 	\bs{I}_3, 7 {\cdot} 10^{-2} \bs{I}_3),  \\
					\bs{Q}_{\mr{ekf}} &{=} \mr{diag} (0.1 {\cdot} 10^{-1} \bs{I}_n,  1 {\cdot} 10^{1} 	\bs{I}_n,  1 {\cdot} 10^{5} \bs{I}_n).
				\end{align}
			\end{subequations}
		\subsection{Simulative Results} \label{ssec:simres}
			A contact-free rectangle trajectory is used to analyze the sensor fusion.
			Maximum speed of~$\| \dot{\bs{x}}_{\mr{t}} \|_{2,\mr{max}} {=} \SI{1.41}{m/s}$ and acceleration of~$\|\ddot{\bs{x}}_{\mr{t}} \|_{2,\mr{max}} {=} \SI{12.44}{m/s^2}$ are reached.
			\begin{figure}[t!]
				\vspace{2.5mm} 
				\centering
				\includegraphics[width=0.9\columnwidth]{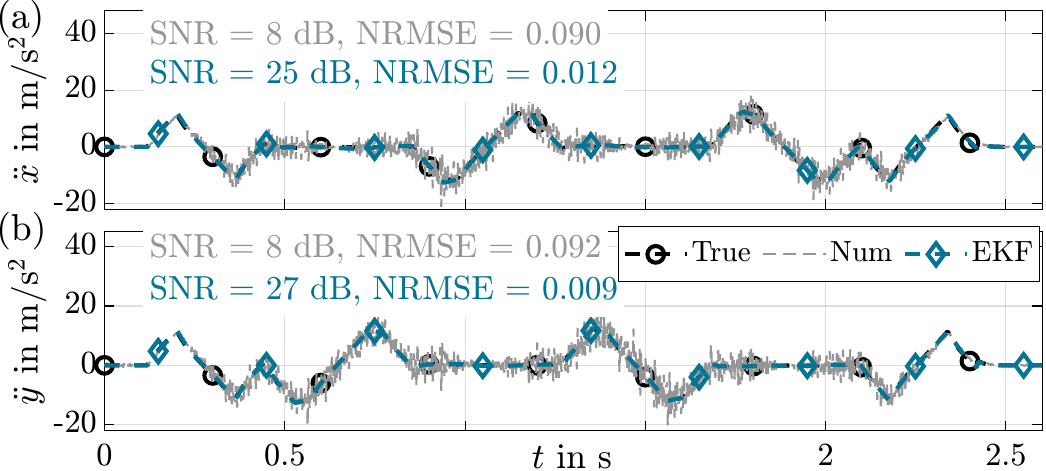}
				\caption{Simulation results with estimated platform acceleration~$\ddot{\bs{x}}_{\mr{E,t}}$ (a) in $x$ and (b) $y$ direction of the EKF compared to double numerical differentiation with low pass filtering~(Num)}
				\label{fig:sim_sf}
				\vspace{-3mm}
			\end{figure}
			Figure~\ref{fig:sim_sf} shows the result of the estimated accelerations.
			With root mean squared errors normalized regard the measurement range~(NRMSEs) of~${<}0.012$ and signal-to-noise ratios~(SNRs) of~${\ge}\SI{25}{\decibel}$, the sensor fusion is accurate, which supports \textit{this work's contribution~\ref{contribution:1IMU} that one IMU is already sufficient for acceleration estimation}.
			For comparison, a causal low-pass filter with a cut-off frequency of~$\SI{20}{\hertz}$ is applied to the two-times numerically differentiated end-effector pose.
			The latter's NRMSEs and SNRs are~$0.09$ and~$\SI{8}{\decibel}$.
			This shows that sensor fusion is necessary for the estimation of~$\ddot{\bs{x}}_{\mr{E}}$.
			
			The direct method is now analyzed using the proposed sensor fusion for different contact scenarios.
			Thresholds are defined to avoid false-positive detection due to modeling inaccuracies.
			If an external force or moment exceeds one of the thresholds, a contact is assumed.
			In the simulation, the thresholds are set to~$\SI{7.5}{\newton}$/$\SI{0.5}{Nm}$ after evaluating the estimation for the previous contact-free rectangle trajectory.
			The detection time is used to evaluate the direct method regarding allowable forces from~\cite{InternationalOrganizationforStandardization.2016}.
			\begin{figure}[b!]
				\vspace{-3mm} 
				\centering
				\includegraphics[width=0.95\columnwidth]{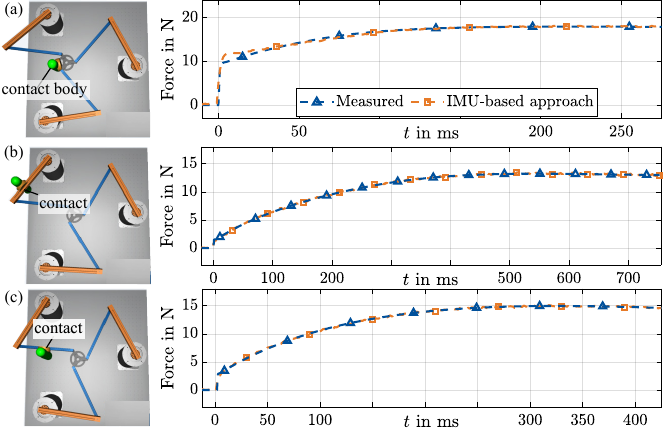}
				\caption{Simulation with measured and estimated external forces at the end-effector platform for~(a) a platform, (b)~first and~(c) second-link collision}
				\label{fig:52_Kontakte_mjx_sim_mp_red}
				\vspace{0mm}
			\end{figure}
			
			Figure~\ref{fig:52_Kontakte_mjx_sim_mp_red}(a) shows the external force of a platform collision with an end-effector speed of~$||\dot{\bs{x}}_{\mr{t,c}}||_2 {=} \SI{0.45}{m/s}$ during contact.
			A contact is detected after $\SI{1}{\milli\second}$ with the direct method, which \textit{underlines contribution~\ref{contribution:Determination_ExtForces}} for simulative results.
			At the same time, false-positive detection is excluded by the selected threshold.
			These observations are analogous to contacts on a serial chain of the PR (see link collisions illustrated in Fig.~\ref{fig:52_Kontakte_mjx_sim_mp_red}(b)--(c)).
			The results of the method are below the lowest allowable contact force for the HRC-relevant upper limbs as defined in ISO/TS 15066.
			There, the allowable upper-limb limits for quasi-static contact are~$110$--$\SI{210}{\newton}$.
			This shows that the use of the direct method in the simulation complies with the limits of safe HRC.
			The simulations show the potential of the direct method so that it can be tested on the test bench.
			%
	
	\section{Validation} \label{sec:validation}
		The test bench~(\ref{ssec:testbench}) and the IMU's integration ~(\ref{sec.imu}) are presented, followed by the experimental results~(\ref{ssec:vsres}).
		\subsection{Experimental Setup} \label{ssec:testbench}
			The experimental studies use the same PR, operated at a sampling rate of~$\SI{1}{\kilo \hertz}$.
			Contact forces are measured by two force-torque sensors\footnote{KMS40 from Weiss Robotics, Mini40 from ATI Industrial Automation}~(FTS).
			EtherCAT and the open-source tool EtherLab are used, which was modified with an external-mode patch and a shared-memory real-time interface.
			Thus, the FTS KMS40 is integrated via a ROS package and the Mini40 via a shared library into the communication with the control system in Matlab/Simulink~\cite{Mohammad.2023}.
		\subsection{IMU Integration and Calibration}\label{sec.imu}
		
			\begin{figure}[t!]
					\vspace{2.5mm} 
					\centering
					\includegraphics[width=0.97\columnwidth]{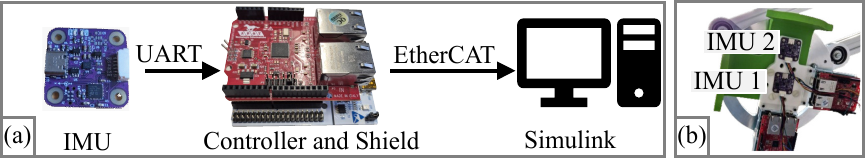}
					\caption{(a) Integration of the IMU in the communication of the test bench and (b) end-effector platform with two IMUs}
					\label{fig.71_Kommunikation_Halterung_red}
					\vspace{-3mm} 
			\end{figure} 
			
			\begin{figure}[b!]
				\vspace{-3mm} 
				\centering
				\includegraphics[width=0.95\columnwidth]{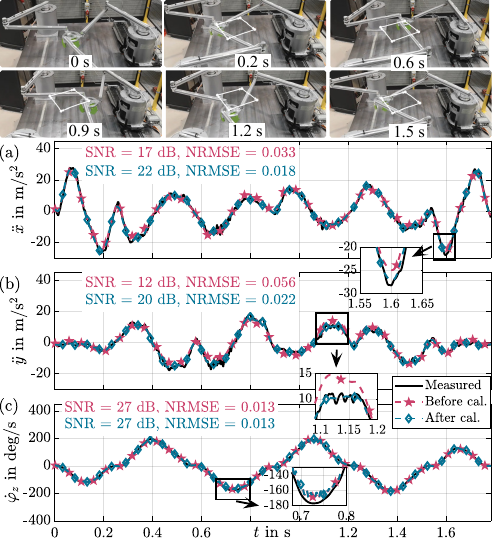}
				\caption{(a)--(b) Measured and modeled accelerations and (c) angular velocity of the second IMU from Fig.~\ref{fig.71_Kommunikation_Halterung_red}(b) before and after calibration}
				\label{fig:72_calibration_imu_dual2}
				\vspace{0mm}
			\end{figure}
			
			\begin{figure}[t!]
				\vspace{2.5mm} 
				\centering
				\includegraphics[width=0.95\columnwidth]{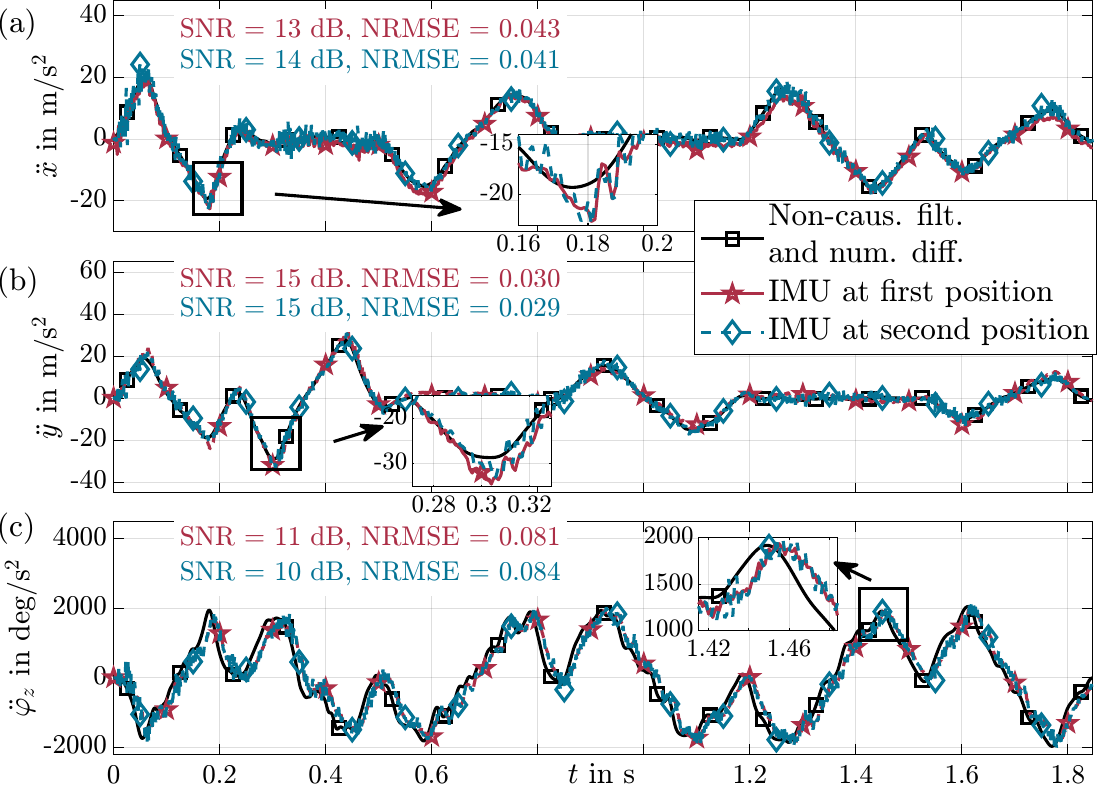}
				\caption{Square trajectory for sensor-fusion tests and estimated~(a)--(b) translational and (c) rotational acceleration in~$\ddot{\bs{x}}_{\mr{E}}$.
						Ground truth is calculated by non-causal and phase-free filtering and numerical differentiation of $\dot{\bs{x}}_\mr{E}$ obtained from the differential kinematics.
						Sensor fusions based on the IMU setups are depicted to evaluate the influence of the IMU's mounting position~$_{(\mr{E})}\bs{p}_{\mr{S}1}$~(IMU$_1$) compared to the displacement~$_{(\mr{E})}\bs{p}_{\mr{S}2}$~(IMU$_2$).
						NRMSEs of the estimation differ by up to~$5\%$ depending on~${_{(\mr{E})}}\bs{p}_{\mr{S}}$ in this example.}
				\label{fig:vierecktraj}
				\vspace{-3mm}
			\end{figure}

			\begin{figure}[b!]
				\vspace{-3mm} 
				\centering
				\includegraphics[width=0.97\columnwidth]{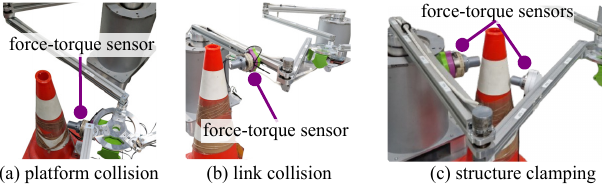}
				\caption{Test bench for collision and clamping contacts}
				\label{fig:vs_aufbau}
				\vspace{0mm}
			\end{figure}
			
			\begin{figure*}[t!]
				\vspace{2.5mm} 
				\centering
				\includegraphics[width=0.95\textwidth]{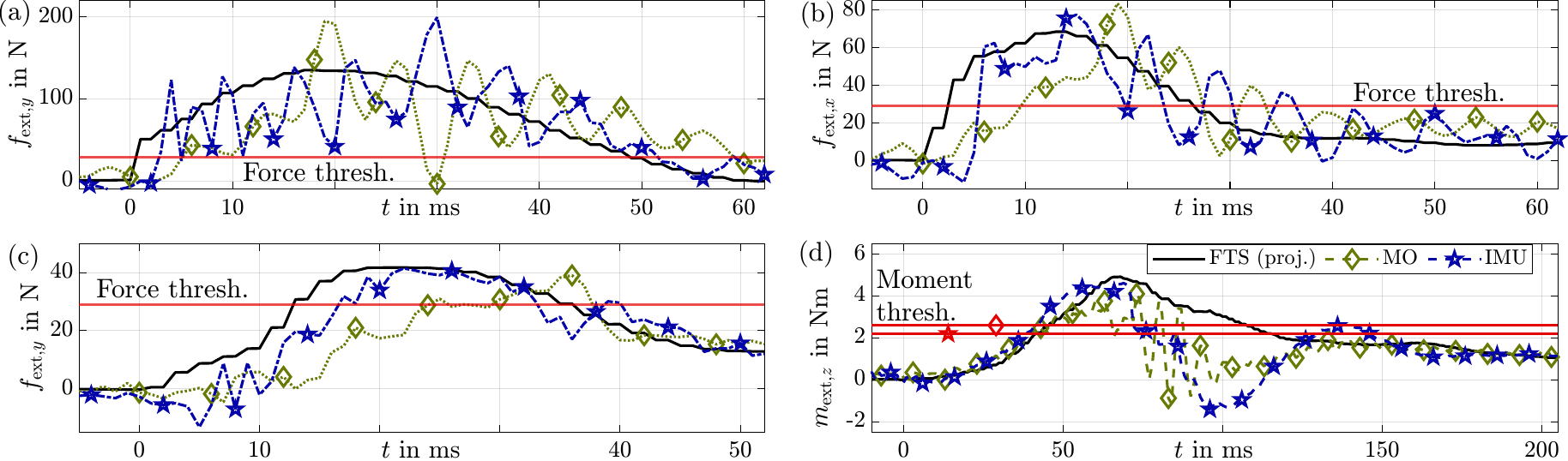}
				\caption{Projected and estimated external forces $f_{\mr{ext},x},f_{\mr{ext},y}$ and moment $m_{\mr{ext},z}$ for comparison of MO and IMU-based direct method during (a)~platform, (b) first and (c) second-link collision and (d) a clamping contact.
					Horizontal lines show the detection thresholds.}
				\label{fig:vs_kontakt}
				\vspace{-3mm}
			\end{figure*}
			
			The IMU is connected to a microcontroller board\footnote{\href{https://www.st.com/en/evaluation-tools/nucleo-f401re.html}{Nucleo-F401RE} from STMicroelectronics} with EtherCAT-Shield\footnote{\href{https://www.bausano.net/en/hardware/easycat.html}{EasyCat-Shield} from Bausano}.
			The communication structure is depicted in Fig.~\ref{fig.71_Kommunikation_Halterung_red}(a).
			The UART (Universal Asynchronous Receiver/Transmitter) protocol is used for sending measurement data from the IMU to the {\micro C}. 
			The EasyCAT shield enables the {\micro C} to transmit measurement values via EtherCAT to the real-time computer of the test bench.
			The consumer-grade\footnote{The total cost for IMU, {\micro C} board and shield is less than 100€.} IMU and {\micro C} with shield are mounted on the end-effector platform, as shown in Fig.~\ref{fig.71_Kommunikation_Halterung_red}(b).
			The IMU-board setting with a data rate of~$\SI{1.6}{\kilo \hertz}$ is chosen. The acceleration and gyroscope measurements are integrated with~$\SI{1}{\kilo \hertz}$ into the communication and have a delay of~$3$--$\SI{4}{\milli\second}$.
			The raw signals include gravitational acceleration, which is compensated for in~(\ref{eq.messa}).
			Internal filtering and data pre-processing are disabled.
			
			The mounting design sets the IMU under the origin of the end-effector frame in the platform's center.
			To test whether an IMU with a larger displacement provides more information, another IMU is attached to the end effector.
			Figure~\ref{fig:72_calibration_imu_dual2} illustrates a contact-free rectangle trajectory and the modeled and measured accelerations and angular rates before and after the experimental identification of the second IMU's mounting position~$_{(\mr{E})}\bs{p}_{\mr{S}2}{=}[75,54,-97]^\mr{T}\mr{mm}$ and the Euler angles~$\bs{\varphi}_{xyz}^\mr{S2,E}{=}[2.2, 3.1, 2.8]^\mr{T}\mr{deg}$, obtained through a particle-swarm optimization of~(\ref{eq:cali}).
			The calibration's positive effect on the reduced NRMSEs and increased SNRs of the modeled translational accelerations are depicted.
			The mounting position and orientation obtained from the first IMU's calibration are~$\bs{\varphi}_{xyz}^\mr{S1,E}{=}[2.8, -5.2, 2.3]^\mr{T}\mr{deg}$ and~$_{(\mr{E})}\bs{p}_{\mr{S}1}{=}[-5,1,-97]^\mr{T}\mr{mm}$.
			
		\subsection{Real-World Results} \label{ssec:vsres}
			
			Figure~\ref{fig:vierecktraj} shows the results of using the EKF with measurement from either the first or the second IMU~(Fig.~\ref{fig.71_Kommunikation_Halterung_red}(b)) for another contact-free rectangle trajectory.
			The estimations are compared with the acceleration obtained through non-causal and phase-free filtering, followed by numerical differentiation of $\dot{\bs{x}}_\mr{E}$, which is determined through the robot's differential kinematics.
			The results show that the states can also be accurately estimated using sensor fusion, which \textit{highlights contribution~\ref{contribution:1IMU}} also on the test bench.
			In addition, the difference between the estimates of the two IMU setups with different~$\bs{p}_{\mr{S}}$ is small, as depicted in Fig.~\ref{fig:vierecktraj}.
			This shows that the IMU mounted with~$||{_{(\mr{E})}}\bs{p}_{\mr{S}2}||_2{>}||{_{(\mr{E})}}\bs{p}_{\mr{S}1}||_2$ at the end effector of the planar PR does not provide additional information compared to the setup with~${_{(\mr{E})}}\bs{p}_{\mr{S}1}$.
			
			The direct method is then compared with the MO regarding the detection time of the collision and clamping contacts, which are shown in Fig.~\ref{fig:vs_aufbau}.
			The maximum velocities and accelerations of the PR's end-effector platform are up to~$\SI{0.8}{\meter / \second}$ and~$\SI{25}{\meter / \second^2}$, which represent high-dynamic contact scenarios. 
			A pylon is chosen as a contact object and fixed to the base to prevent its displacement. 
			As soon as contact is detected, the robot's movement is stopped.
			According to Kalman's observability criterion, the linearized system is locally observable in the following trajectories.

			Table~\ref{table:kontakt_platform} shows the detection time of MO- and IMU-based contact detection for collisions and clamping contacts.
			Generalized-force thresholds for binary contact detection are chosen to correspond with a safety factor of~$2$ to the maximum estimated external forces and moments during the contact-free trajectory in Fig.~\ref{fig:vierecktraj}.
			For a fair comparison, the observer gain~$k_{\mr{o},i}$ of the MO is varied in the range of~$20$--$500\frac{{1}}{\mr{s}}$, which led to force and moment thresholds of~$13$--$\SI{108}{\newton}$ and~$0.8$--$\SI{9.2}{Nm}$ with regard to the estimation in operational-space coordinates.
			The lowest observer gain is chosen for the initiation of the contact reaction during the experiments, ensuring that thresholds of the higher gains are reached.
			The direct method's thresholds are~$\SI{29}{\newton}$/$\SI{2.2}{Nm}$ and differ in comparison to the simulation in Sec.~\ref{ssec:simres} due to modeling inaccuracies in the dynamics.
			Despite the increased thresholds, these are below the permitted limits in the ISO/TS~15066 ($110$--$\SI{420}{\newton}$ for transient and quasi-static contacts).
			The results show that the \textit{direct approach detects all contacts within milliseconds and faster than the MO}, which shows the direct method's superior performance and \textit{proves contribution~\ref{contribution:Determination_ExtForces} and~\ref{contribution:Contact_Detection}}.
			\begin{table}[t!]
				\vspace{0mm}
				\centering
				\caption{Detection duration $\Delta t_\mr{det}$ in ms for the contact experiments in Fig.~\ref{fig:vs_aufbau} with different observer gains of the MO.}
				\vspace{-1.5mm}
				\begin{tabular}{l||c|c|c|c|c||c}
					Detection approach& \multicolumn{5}{c}{MO with $k_{\mr{o},i}$ in $1/\mr{s}$}& IMU \\
					parameterization& $20$ & $100$& $135$ &$200$&$500$& (\ref{eq:ekf_setting}) \\
					\hline
					Threshold in N & 13 & 22 & 29&43 &108 & 29  \\
					\hline 
					Threshold in Nm & 0.8 &  2 & 2.6&3.8 &9.2 & 2.2  \\
					\hline 
					Platform collision&12 &6&6 &6 &6 &\textbf{3}\\ 
					\hline 
					First-link collision&14 &10 &10 &11 &- &\textbf{6}\\ 
					\hline 
					Second-link collision&34 &24 &25 &- &- &\textbf{17}\\ 
					\hline 
					Chain clamping&41 &41 &47 &- &- &\textbf{39}\\ 
				\end{tabular}
				\label{table:kontakt_platform}
				\vspace{-3.5mm}
			\end{table}
			
			Figure~\ref{fig:vs_kontakt} and Table~\ref{table:vs_kontakt} show more detailed results of the contact scenarios for choosing~$k_{\mr{o},i}{=}135\frac{{1}}{\mr{s}}$ to have the same force threshold~$\SI{29}{\newton}$ as the IMU-based approach.
			The measured and estimated data are respectively expressed in the operational-space coordinate with the largest effect caused by the contact.
			\begin{table}[b!]
				\vspace{-3.5mm}
				\centering
				\caption{Improvement of the detection method}
				\vspace{-1.5mm}
				\begin{tabular}{l||c|c|c|c}
					& Platform& 1st link & 2nd link& Clamping \\
					Experim. in Fig.~\ref{fig:vs_kontakt}&(a)& (b)& (c)& (d) \\\hline \hline
					%
					%
					%
					$\Delta t_\mr{det,MO}$ in ms & 6 & 10 & 25 & 47 \\\hline
					%
					%
					$\Delta t_\mr{det,IMU}$ in ms &  3 &  6 &  17 &  39\\\hline
					%
					%
					\textbf{Reduction in \%} & \textbf{50} & \textbf{40} & \textbf{32} & \textbf{17} \\\hline
				\end{tabular}
				\label{table:vs_kontakt}
				\vspace{0mm}
			\end{table}
			The chain contacts' results~(b)--(d) show the external forces, respectively moment measured by the FTS and projected onto the end effector by the corresponding contact Jacobian~\cite{Mohammad.2023}.
			Since both the direct method and the MO estimate the external force acting on the platform, this is the comparative value for the quantitative categorization of the methods.
				\subsubsection{Collision Results}
					Figure~\ref{fig:vs_kontakt}(a)--(c) shows the measured and estimated contact forces during platform, first- and second-link contacts.
					The respective detection duration~$\Delta t_{\mr{det},i}$ is given in Table~\ref{table:vs_kontakt}.
					Compared to the MO, the direct method reduces the detection time for collisions from~$6$--$\SI{25}{\milli \second}$ to $3$--$\SI{17}{\milli \second}$ and for clamping scenarios from~$\SI{47}{\milli \second}$ to~$\SI{39}{\milli \second}$, which \textit{underlines contributions~\ref{contribution:Determination_ExtForces} and~\ref{contribution:Contact_Detection} of this work}.
					However, the time series in Fig.~\ref{fig:vs_kontakt}(a) show strong oscillations, which are caused by the sudden jerk in the event of contact.
					These vibrations occur with smaller amplitude and frequency during link collisions in Fig.~\ref{fig:vs_kontakt}(b)--(c), which suggests that the vibrations induced by the contact are damped by the structure or the platform collision especially excites unmodeled effects such as structural elasticity and joint clearances.
					Both link collisions show that the direct method captures the extreme values of the projection faster than the MO due to the MO's inherent time delay.
				\subsubsection{Clamping Results}
					Figure~\ref{fig:vs_kontakt}(d) shows the moments estimated on the platform by the MO and direct approach, as well as projected measurements according to~(\ref{eq:jacobian_clamping}).
					The IMU-based approach requires~$\SI{39}{\milli \second}$ compared to the MO with~$\SI{47}{\milli \second}$, although the rotational platform accelerations are not measured and only estimated in the algorithm.
					The abrupt change after~$t{=}\SI{50}{\milli \second}$ is attributed to the reaction.
					
				To summarize, the direct method (i) estimates dynamic changes in the external force and (ii) detects contacts with up to $50\%$ shorter detection duration~(see Table~\ref{table:vs_kontakt}).
				The probability of false-positive detection is reduced by parameterizing the thresholds with dynamic, contact-free trajectories.
	\section{Conclusion} \label{sec:conlusions}
		This work presents a sensor fusion of inertial measurement units (IMUs) and encoders to estimate platform accelerations.
		The use of a parallel robot~(PR) allows mounting only one IMU on the end-effector platform to estimate the acceleration of the robot's minimal coordinates.
		Based on an extended Kalman filter and an identified dynamics model, external forces are estimated to detect contacts within $3$--$\SI{39}{\milli\second}$.
		A shorter detection time also reduces contact forces if, for example, a retraction is performed.
		The up to~$50\%$ faster detection highlights the potential of IMU-based contact detection, which favors the commercial utilization of PRs operating at high speeds in human-robot collaboration.
		Future research will focus on the application of the methods to spatial PRs, \highlightred{with an analysis of the observability and their nonlinearity that could motivate the use of an unscented Kalman filter.}
		Also, a comparison with recent approaches is planned, such as data-driven methods with physically modeled features or more complex model-based observers.
	\addtolength{\textheight}{0cm} 
	
	

	\bibliographystyle{IEEEtran}
	\bibliography{literatur}	
\end{document}